# A Novel Kinesthetic Haptic Feedback Device Driven by Soft Electrohydraulic Actuators

Dannuo Li, Quan Xiong, Xuanyi Zhou, and Raye Chen-Hua Yeow, *Member, IEEE*

*Abstract*—Developing kinesthetic haptic devices with advanced haptic rendering capabilities is challenging due to the limitations on driving mechanisms. In this study, we introduce a novel soft electrohydraulic actuator and develop a kinesthetic haptic device utilizing it as the driving unit. We established a mathematical model and conducted testing experiments to demonstrate the device's ability to stably output controllable feedback force. Our experiments also demonstrates that this device exhibits fast response characteristics. By utilizing the easily controllable nature of the soft electrohydraulic actuator, we were able to achieve high-resolution controllable feedback force output. Furthermore, by modulating the waveform of the driving high-voltage, the device acquired the capability to render variable-frequency haptic vibration without adding any extra vibration actuator. Using this kinesthetic haptic device, we built a teleoperated robotic system, showcasing the device's potential application as a haptic force feedback system in the field of robotics.

*Key words*—Soft actuators, electrohydraulic actuators, kinesthetic haptic, human-robot interaction, teleoperated robotic system.

## 1. Introduction

Kinesthetic haptic devices provide physical feedback to users by applying forces, vibrations, or motions to human finger [1][2]. These devices simulate the sensation of interacting with physical objects, allowing users to perceive weight, texture, resistance, and other physical properties [3][4][5][6]. Over the past decades, such devices have attracted significant interest in both industry and academia for applications in medical simulations, virtual reality (VR), remote control systems (telerobotics) and gaming [7][8][9]. Kinesthetic haptic devices are mainly driven by motors. For example, Pierce *et al*. [10] demonstrated a wearable device that uses a geared DC motor to output reverse torque for rendering feedback force. Cempini *et al*. [11] proposed a hand exoskeleton that generates resistance for haptic kinesthetic feedback by using a geared DC motor-capstan-tendon mechanism. Kinesthetic haptic devices driven by geared motors offer easy control and stable output capabilities. However, the presence of the gearbox in the motor results in slow response times and low back drivability, which limits the agility and safety of haptic devices.

To circumvent these challengers, alternative actuation methods was explored. Zubrycki *et al*. [12] developed a vacuum based soft pneumatic actuator (SPA) that was palm-mounted, when actuated a jamming force was generated. Alternatively, positive pressure can be used to generate braking forces-Jadhav *et al* [13]. SPAs provide haptic devices with safe and supple interaction properties. However, precise control of SPAs is very challenging. Moreover, SPAs together with-it power system can be considered bulky, noisy and inconvenient.

Recently, a new type of soft actuator based on electrostatic attraction (EA) called electrohydraulic actuators, as known as hydraulically amplified self-healing electrostatic (HASEL) actuators [14][15][16], was explored for haptics [17][18]. When actuated HASEL actuators are capable of displacement and force output by regulating applied high voltages. Furthermore, they have a simple fabrication process and excellent performance such as having rapid response and a high power to weight ratio. However, most existing HASEL actuators are driven by constant DC voltage, resulting in unstable force and displacement output [16][19][20][21][22][23][24][25]. And this characteristic limits the research of HASEL actuators in the field of haptic to cutaneous haptic applications [17]. In our previous work, we developed new polyimide-based HASEL actuators driven by low-frequency AC square wave high-voltage to achieve stable displacement output [26].

In this work, we delve further in the application utilizing our prior HASEL actuator. We developed a novel kinematic haptic feedback device, which is stable, fast and has controllable force feedback during interaction with users. The device, mounted on a horizontal tabletop, enables haptic force feedback of a user's fingers imitating the pinching of an object. This is achievable by linear-stacked HASEL actuators positioned on both sides of the device allowing the output controllable force. The force generated by the actuators is transmitted to the user's fingers through two symmetrically distributed connecting rods within the kinesthetic haptic device. In addition, we devised a mathematical model of the relationship between the squeeze displacement and force output of the HASEL actuator under different driving voltages. This mathematical model was then validated through experiments. Subsequently, we designed a force amplification and transmission mechanism to be used in the device to realize the amplification of the feedback force of the HASEL actuators. Next, we established a statics model of this force amplification and transmission mechanism, demonstrating experimentally that this mechanism effectively amplifies and transfers the HASEL actuator's output force to the user's

This research is supported by A*STAR Industry Alignment Fund - Pre-Positioning (A20H8A0241). (Corresponding author: Chen-Hua Yeow, Quan Xiong). Dannuo Li and Quan Xiong contributed equally.

Dannuo Li, Quan Xiong, Xuanyi Zhou and Chen-Hua Yeow are with the Department of Biomedical Engineering and Advanced Robotics Centre, National University of Singapore, Singapore, 117583 Singapore (e-mail: e0978500@u.nus.edu, quan.xiong@u.nus.edu, e0978497@u.nus.edu, rayeow@nus.edu.sg).

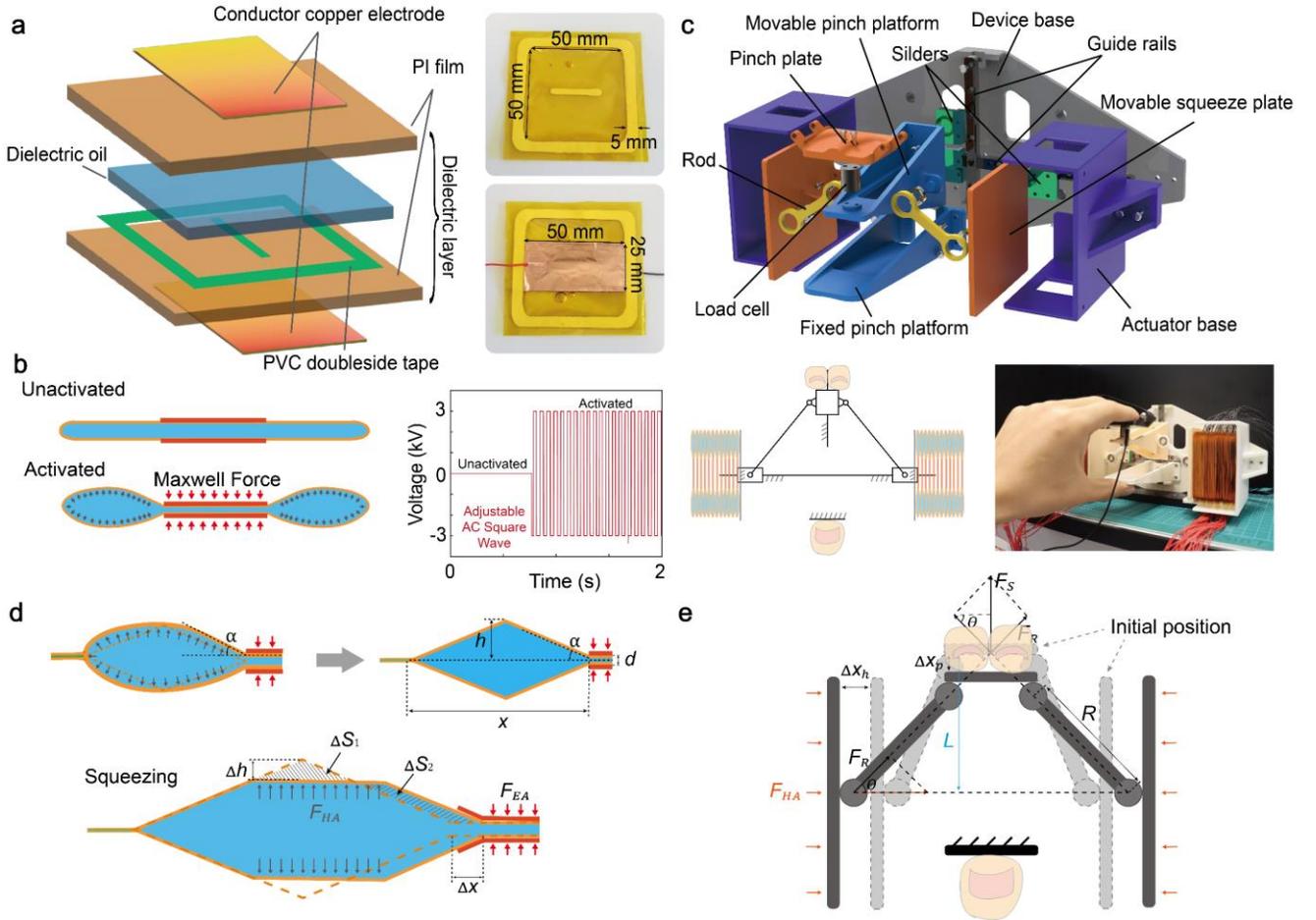

**Fig. 1.** (a) The structure of HASEL actuator and actual picture. (b) The principle of HASEL actuators. (c) The structure of force amplification and transmission mechanism, schematic diagram of the mechanism and the actual kinesthetic haptic device. (d) The mathematical model of HASEL actuator. (e) The model of mechanism.

fingers. We also demonstrated the device's the ability to generate effective feedback force with a short response time in experimental tests. By modulating the AC activation voltage waveform, we enabled the device to produce variable-frequency haptic vibrations alongside the feedback force, without adding any extra vibration actuator. Finally, we developed a teleoperated robotic system based on this kinesthetic haptic device, which can simulate the teleoperation of the pinching motion with real-time force feedback.

## 2. RESULTS

### 2.1 HASEL Actuator Design

Polyimide is a dielectric material commonly used in the microelectronics industry for its high temperature resistance, good thermal conductivity, high mechanical strength and relative permittivity (3.4). The composition of the HASEL actuator shown in Fig. 1a. Two PI films of 25 μm thickness were used as the insulating liquid bladder material in the HASEL actuator. PVC doubleside tape was used to bond the two layers of PI film under high temperature heating (200 °C). 25 ml of dielectric oil was used as the liquid medium inside the bladder. Two conductor copper electrodes were affixed to the actuator in the centered position on the top and bottom sides. The geometrical para meters of each part as well as the physical object are shown in Fig. 1a. More details can be found in Sec 5.1 and Ref [26].

The driving principle of the HASEL actuator is shown in Fig. 1b. When AC square wave voltage is deactivated, the actuator is in unactivated state and the dielectric liquid is uniformly distributed in the liquid bladder. When AC square wave voltage is applied to the electrodes, the two electrodes are attracted towards each other by the Maxwell force. This causes the dielectric liquid within electrode plates to be squeezed to the sides of the liquid bladder thus causing deformation to output displacement and generating output force.

### 2.2 Force Amplification and Transmission Mechanism Design

To transmit and amplify this feedback force, a force amplification and transmission mechanism for the kinesthetic haptic device was designed and fabricated (see Fig. 1c). The base of the mechanism was fixed on top a table with its bracket on its back, this is to allow the user to interact with the device in a natural horizontal position. The front of the base was fitted with two metal guide rails (MGN7 200 mm, DEBOERKEJI) in a crisscross pattern. Two actuator bases were mounted symmetrically at the left and right ends of the device base. The pinch platform for the user's thumb was mounted in the bottom position of the base. The movable pinch platform was mounted on two side-by-side sliders (MGN7C) on vertically guide rail. A load cell (ZNLBM-X) was mounted at its end to monitor the feedback force felt by the user during the interaction. Pinch plate for the user's index and middle fingers was mounted on the upper of the load cell. Two square movable squeeze plates (50×50mm) were mounted symmetrically on two sliders on horizontal guide rail. The squeeze plates were connected to the movable pinch platform by means of rolling bearings and connecting rods (length: 35mm). The entire rigid mechanism, with the exception of the guide rails, sliders and rolling bearings, was manufactured using 3D printing (Ultimaker, ABS, 230 ℃).

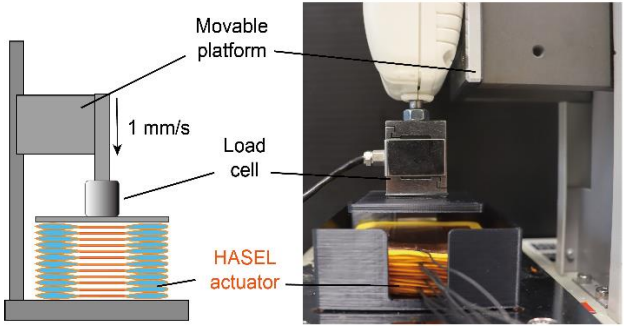

**Fig. 2.** The mathematical model validation experimental setup.

The assembled kinesthetic haptic device shown in Fig. 1c has linearly stacked HASEL actuators (30per side) placed in both actuator base. When the user pinches on the pinch plate, it drives the movable squeeze plates via the connecting rods to squeeze the activated actuators. The force generated by the actuators onto the movable squeeze plates is transferred to the user's fingers through the connecting rods to complete the force feedback.

*2.3 Modeling of Device Driving and Output Capabilities*

The stable and controllable feedback force generated by the HASEL actuators during compression, can be model to provide a better understanding of the system and optimization. We developed a mathematical model of the squeeze displacement-feedback force for the single HASEL actuator. In addition, we present a kinematic model of the force amplification and transmission mechanism. The corresponding model validation experiments will be included in Sec. IV.

In an ideal case, when the HASEL actuator is active and not squeezed, we can ignore the residual dielectric liquid between the two electrodes and assume that all of the dielectric liquid is symmetrically squeezed into the liquid bladder regions on the left and right sides of the HASEL actuator. The liquid bladder is deformed at a small angle $\alpha$ by the combined action of the edge PVC double-sided adhesive and the high mechanical strength PI film, thus we can approximate its cross-sectional shape as a centrosymmetric rhombus (Fig. 1d). Half the height of the vertical expansion of the dielectric bladder $h$ can be calculated as follows,

$$h = \frac{V}{2xL}, \tag{1}$$

where $V$ is the total volume of dielectric oil (2.5ml), $x$ is the width of the unilateral liquid bladder (12.5mm), and $L$ is the length of the liquid bladder (50mm). The angle $\alpha$ can be calculated as follows,

$$\alpha = \arctan\frac{2h}{x}, \tag{2}$$

when squeezed in the vertical direction, we approximate the inner side of the rhombic liquid bladder will keep the deformation angle $\alpha$ to advance toward the electrode side. Since the liquid is incompressible, the area reduced in the vertical direction $\Delta S_1$ and the area increased in the horizontal direction $\Delta S_2$ can be given as:

$$\Delta S_1 = \Delta S_2 \tag{3}$$

The $\Delta S_1$ can be calculated as follows,

$$\Delta S_1 = \frac{\Delta h^2}{\tan \alpha} \tag{4}$$

The $\Delta S_2$ can be calculated as follows,

$$\Delta S_2 = \Delta x \cdot (h - \Delta h), \tag{5}$$

where $\Delta x$ is the horizontal displacement of the liquid bladder advancing toward the electrodes side.

The area of the squeezed surface of the HASEL actuator can be calculated as follows,

$$S_{HA} = \left(\frac{2\Delta h}{\tan \alpha} + \Delta x\right) \cdot L, \tag{6}$$

The electrostatic adhesion (EA) force between the two electrodes, also as known as the Maxwell force, can be calculated based on the parallel capacitor model as follows,

$$F_{EA} = \frac{1}{2}\frac{\varepsilon_r \varepsilon_0 A u_0(t)^2}{d^2}, \tag{7}$$

where $\varepsilon_0$ is the vacuum permittivity, $\varepsilon_r$ is the relative permittivity of the dielectric layer, $A$ is the overlap area of the two electrodes, and $d$ is the thickness of the dielectric layer between two electrodes.

The pressure of the liquid in the dielectric bladder can be calculated as follows,

$$P = \frac{F_{EA}}{A} = \frac{1}{2}\frac{\varepsilon_r \varepsilon_0 u_0(t)^2}{d^2}, \tag{8}$$

the feedback force in the vertical direction generated by the HASEL actuator can be calculated as follows,

$$F_{HA} = 2PS_{HA}, \tag{9}$$

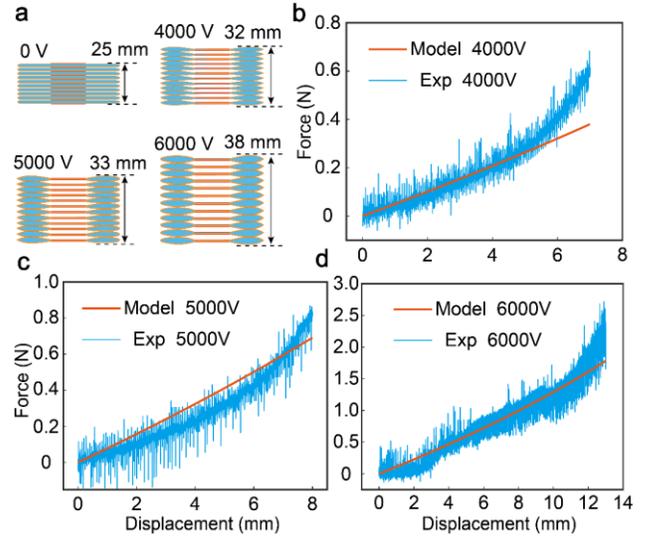

**Fig. 3.** (a) The HASEL actuators stack height at different activation voltages. (b, c, d) Feedback force-squeeze displacement relationship.

TABLE I
PRE-EXPERIMENTS FOR SOLVING MODEL PARAMETERS. D(MM) AND F(N)
REPRESENT THE SQUEEZE DISPLACEMENT AND FEEDBACK FORCE

|  | D1 = 0.125 | D2 = 0.250 | D3 = 0.375 |
|---|---|---|---|
| F1 | 0.2785 | 0.5260 | 0.8552 |
| F2 | 0.2830 | 0.5250 | 0.8573 |
| F3 | 0.2750 | 0.5273 | 0.8573 |
| Average (N) | 0.2788 | 0.5261 | 0.8566 |

By associating (1) to (9) above, the relationship between the squeezed displacement $\Delta h$ and the feedback force $F_{HA}$ is obtained. However, an unknown volume of dielectric liquid exists between the overlapping regions of the two electrodes. The thickness of this part of the liquid is associated with the $\varepsilon_r$ and $d$ parameters in Eq. 7 and Eq. 8.

Pre-experiments were conducted to obtain the values of mixing parameter $K$:

$$K = \frac{1}{2}\frac{\varepsilon_r \varepsilon_0}{d^2} \quad (10)$$

We repeatedly measured the feedback force output from a linear stack of three HASEL actuators at three constant squeeze displacement points (see Table I). The data obtained from the pre-experiment were averaged and brought into the above mathematical model equation system and to solve for the mixing parameter $K$ (9.828e-5).

Thus, we have completed the modeling of the relationship between the squeeze displacement and the feedback force of the HASEL actuator during different amplitudes of voltages activation.

A simple illustration of the force amplification and transmission mechanism is shown in Fig. 1e. The displacement of the left and right squeeze plates relative to their initial positions at a certain point in the user's pinching process can be calculated by:

$$\Delta x_h = \sqrt{R^2 - (L - \Delta x_P)^2} - \sqrt{R^2 - L^2}, \quad (12)$$

where $R$ is the length of the connecting rod, $L$ is the vertical distance between the pinch plate initial position and the horizontal line at the lower end point of the connecting rod, and $\Delta x_P$ is the displacement of the user's vertical pinch. Knowing the total squeeze displacement $\Delta x_h$, the feedback force $F_{HA}$ generated by the linearly stacked HASEL actuators can be obtained by combining the mathematical model built in Sec. III.A.

At this point, the angle between the connecting rod and the horizontal direction can be calculated as follows,

$$\theta = arcsin\left(\frac{L - x_P}{R}\right), \quad (13)$$

the component force of the feedback force in the direction of the connecting rod can be calculated as follows,

$$F_R = F_{HA} \sin\theta, \quad (14)$$

synthesize the force transmitted on the two connecting rods into the vertical direction as the feedback force felt by the user's finger:

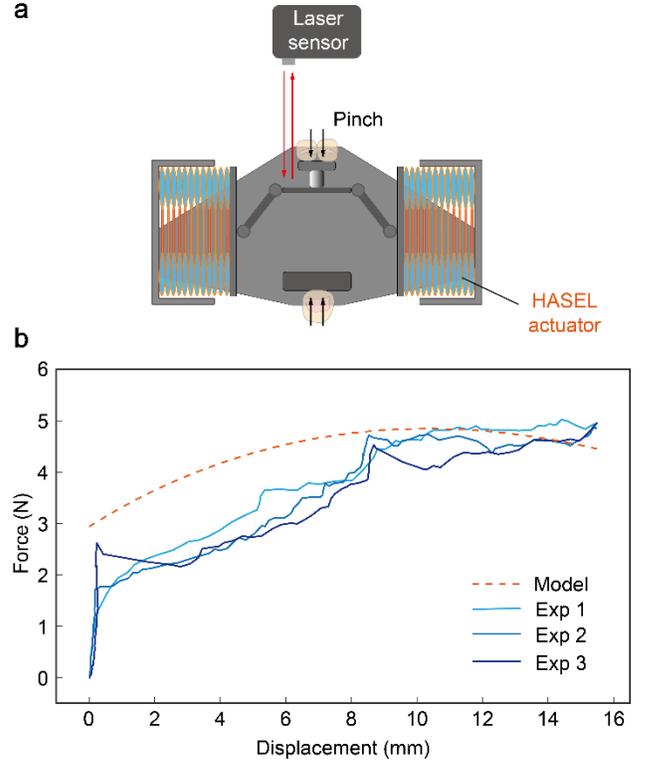

**Fig. 4.** Device feedback force output experiment. (a) Experiment setup. (b) Maximum feedback force output during pinching motion at 6 kV AC voltages.

$$F_s = 2F_R \sin\theta, \quad (15)$$

*2.4 HASEL Actuators Output Force Validation*

To measure the output force by activated HASEL actuators when squeezed, we fixed 10 linearly stacked HASEL actuators on a test machine shown in Fig. 2. The lower end of the movable platform in the test machine was connected to a load cell, attached to both square squeeze plate (50×50mm) on each side. We validated the output force by recording the data acquired from the load cell as the platform moved downwards at 1mm/sec.

The height of the HASEL actuator stack under the activation of an AC square wave high voltage at different amplitudes and a frequency of 20 Hz is shown in Fig. 3a, illustrating the effective squeeze displacement at different activation voltages. We kept the applied voltage below 6000 V as voltage higher than 7000 V would cause an electrical breakdown.

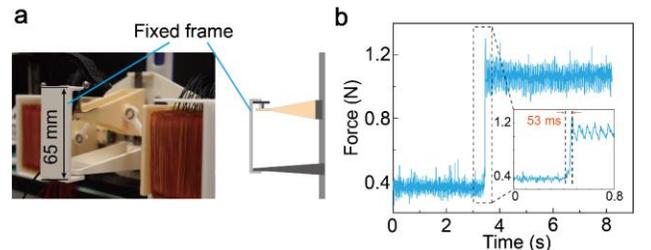

**Fig. 5.** (a) Experimental setup for constant pinch displacement. (b) Device response time.

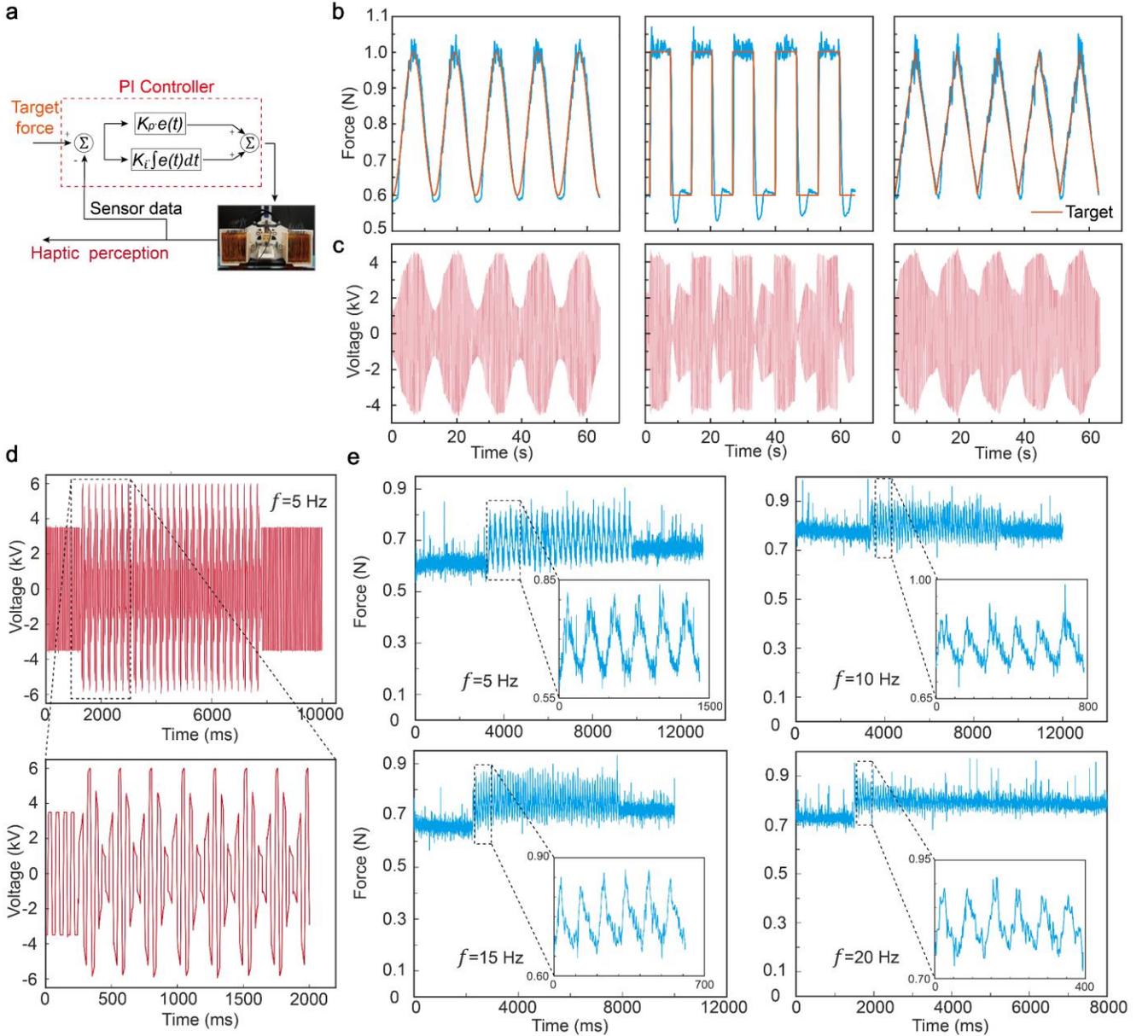

**Fig. 6.** (a) The feedback control diagram. (b) The feedback force output with feedback control. The device following targets waves. (c) The actual applied voltages. (d) The drive voltage waveform with a 5Hz sine wave is superimposed. (e) Haptic vibrations produced by superimposing sine waves of different frequencies.

As shown in Fig. 3b,c,d, the experimental data of output force closely follows the calculation based on model built on Sec III.A. This validates our mathematical model of squeeze displacement-feedback force for HASEL actuators.

*2.5 Device Maximum Force Output*

To measure the maximum force of the device output, we installed a laser sensor above the device to measure the displacement change of the movable pinch platform (see Fig. 4a). At 6000 V voltage activation, three pinch experiments were conducted. As shown in Fig. 4b, the device exhibits an effective pinch stroke of 15 mm, with the maximum feedback force ranging approximately from 2 N to 5N. The sudden change in output force at the pinch displacement of 0 mm is because the HASEL actuator stacks loaded on the left and right sides of the device are in the state of squeeze before the start of the pinch movement.

The experimental data of output force shares the same tendency with calculation based on kinematic model built on Sec III.B. The experimental results exhibit a significant deviation from the target in the first half of the pinching movement, possibly due to the dynamic nature of pinching rather than static equilibrium. This discrepancy is unacceptable for the haptic feedback device, as the user can perceive the force difference. Therefore, closed-loop control is necessary to ensure consistent output.

*2.6 Device Response Time*

We built the experimental setup shown in Fig. 5a. The distance between the upper and lower pinch platforms was fixed by a 65 mm 3D-printed rigid frame, ensuring that the displacement of the pinch movement remained constant during the experiment. Under open-loop control, we measured the output force jump curve while varying the input voltage amplitude (3 kV-6 kV). As shown in Fig. 5b, the device's output responds rapidly, with a response time of approximately 53ms. This high responsiveness enhances the device's ability to render kinesthetic haptic effectively.

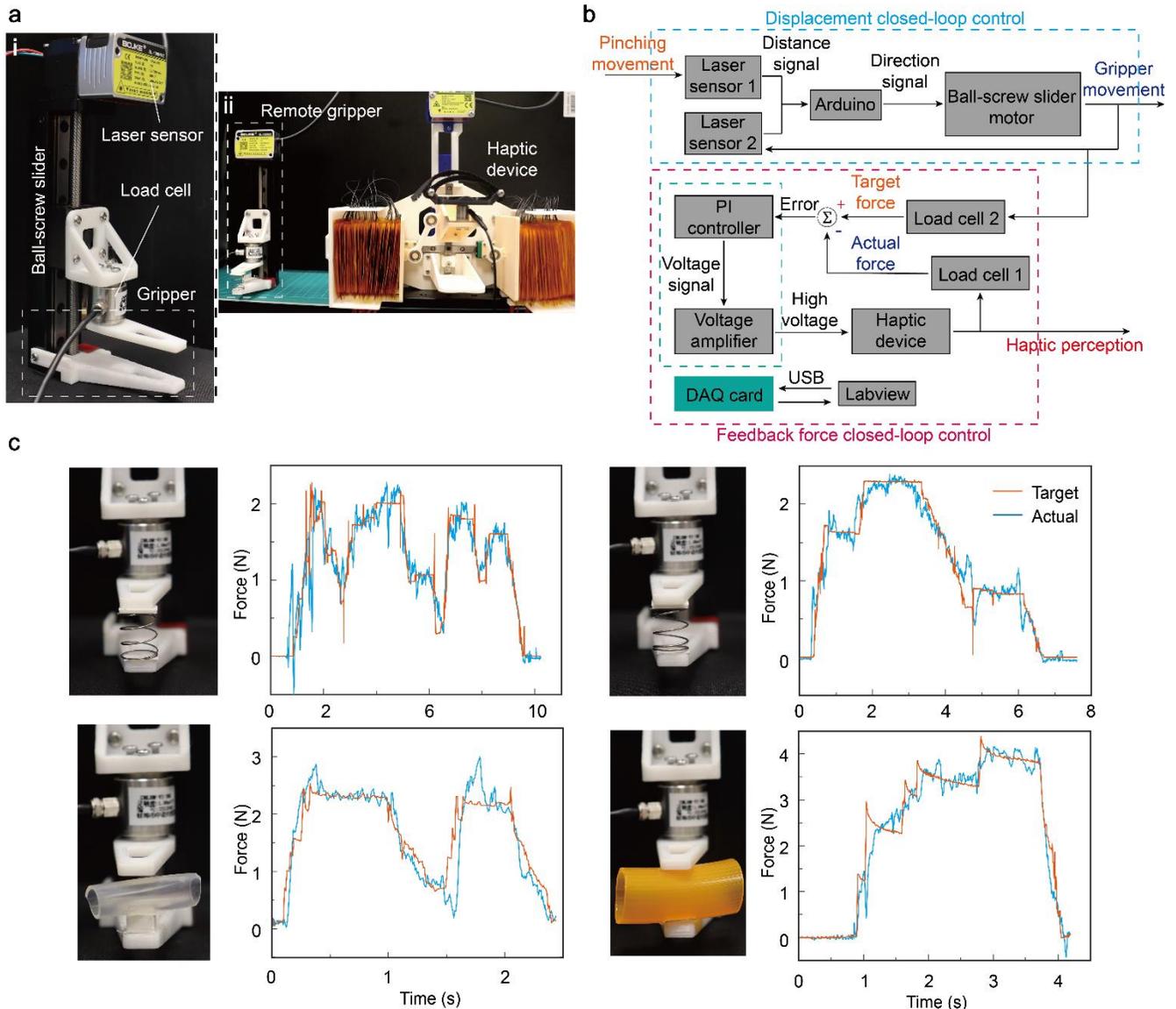

**Fig. 7.** (a.i) Slave gripper robot. (a.ii) Teleoperated robotic system. (b) Control flow diagram of teleoperated robotic system. (c) The feedback force felt by the user and the actual force on the gripper during the teleoperation of the grasping of four different softness objects.

### 2.7 Feedback Control Based Fine Force Control

To achieve fine force output regulation, we implemented a feedback control on our device using a Proportional-Integral (PI) controller. The control flow is illustrated in Fig. 6a. The control cycle was set to 1 ms on the LabVIEW platform, and the controller parameters $K_p$ and $K_i$ were manually adjusted to 0.75 and 0.035, respectively. The same experimental setup was used to ensure constant pinch displacement (see Fig. 5a). We then implemented the feedback control to follow the target force waves (0.08 Hz sine waves, square waves and triangle waves). The force and applied voltages were recorded in Fig.

6b and Fig. 6c. The actual feedback force slightly lags behind the target at the falling edges, likely due to the HASEL actuator generating only unidirectional (outward) force.

### 2.8 Haptic Vibration

The HASEL actuators are driven by AC voltage. In our previous work [26], when using low-frequency AC voltage drives, the HASEL actuators exhibited vibration. Normally vibration affects the performance of the actuator, but if controlled vibration can be achieved, this can be an advantage.

Using the same experimental setup shown in Fig. 5a, we can enable the output of variable frequency haptic vibration, by superimposing a sinusoidal AC voltage of different frequencies on a high-frequency AC square wave as the activation voltage. Fig. 6d illustrates the morphology of the voltage waveform after superimposing a sinusoidal AC wave (5 Hz, 2.5 kV) onto the original AC square wave (20 Hz, 3.5 kV). As shown in Fig. 6e, the device outputs haptic vibration of different frequencies when superimposing sine waves of different frequencies. When the superimposed sine wave frequencies are 5 Hz, 10 Hz, 15 Hz and 20 Hz, the average of the actual measured vibration frequencies are approximately 4.37 Hz, 8.75 Hz, 9.80Hz and 15.15 Hz, respectively. Hence, the amplitude of the vibration is within the range of 0.15N to 0.2N, demonstrating the ability of the device to achieve controlled haptic vibration output.

*2.9 Teleoperation Demonstration*

A teleoperated robot, controlled through remote technology, enables users to perform complex tasks in hazardous or inaccessible environments [28][28]. Force feedback can significantly enhance the precision of the user's maneuvers and provide a better understanding of the physical properties of the remote environment [29][30][31][32][33][34].

To demonstrate the performance of the kinesthetic haptic device in real-time force feedback applications, we designed and built a gripper teleoperated robotic system, as shown in Fig. 7a. This system consists of the remote gripper as the slave robot and the kinesthetic haptic device as the operating side (see Fig. 7a.ii). The structure of the remote gripper is shown in Fig. 7a.i. One end of the 3D-printed rigid gripper was affixed to the end of the ball-screw slider, while the other end, was equipped with a load cell, which was mounted on the movable platform of the ball-screw slider. At the opposite end of the ball-screw slider, the distance laser sensor was mounted above the driven motor to measure the displacement data of the movable platform. The operating side consists of the kinesthetic haptic device and a distance laser sensor above it for measuring pinch displacement.

Fig. 7b illustrated the teleoperation workflow and the process of generating real-time force feedback. When the user interacts with the kinesthetic haptic device, the Arduino microcontroller reads the data from the laser sensor on top of the device in real time (sampling rate: 1000) and compares it with the initial position data to get the direction information of the displacement. Based on this direction information, the microcontroller sends the drive signal to the motor. It then uses the real-time data read by the laser sensor at the gripper side as feedback to control the gripper, ensuring it generates displacement in the same direction and of the same distance. After the gripper contacts with the object, the force read by the load cell on the gripper is used as the target force, while the force read by the load cell in the kinesthetic haptic device is used as the actual force. Through the action of the Proportional-Integral controller, closed-loop control is achieved to output real-time feedback force.

Grasp perception experiments with physical objects were conducted using this teleoperated robotic system. Compression springs of two wire diameters (0.5mm, 0.6mm) and hoses of two different hardnesses (soft and semi-rigid) were grasped (see video V1). The real-time feedback force felt by the user's fingers during the grasping process shown in Fig. 7c.

## 3. Discussion

We developed a kinesthetic haptic device based on a novel soft electrohydraulic actuator (HASEL actuator). Our kinesthetic haptic device offers significant advantages in responsiveness compared to motor-driven haptic feedback devices and does not require a complex and bulky drive source when compared to pneumatic soft actuators. Moreover, it achieves precise feedback force real-time control through simple closed-loop control during interaction. The HASEL actuator, used as the device's drive unit, has significant advantages over conventional drives in terms of low cost (0.031 $ each) and easy fabrication. Furthermore, in the event of the failure of a HASEL actuator, maintenance can be easily accomplished by simply replacing the defective actuator.

When using low-frequency, high-voltage AC square wave to drive the HASEL actuator, we observed that the actuator would vibrate (see video V2). This vibration frequency changed corresponds to the square wave frequency. When we activate the HASEL actuator with a high-frequency (> 20Hz), high-voltage square wave, the vibration phenomenon is eliminated. Consequently, when the user interacts with the device, they feel a stable feedback force without any vibration. Based on this phenomenon, we achieved variable frequency haptic vibrations output from the device while maintaining stable feedback force output by superimposing a low-frequency sine wave onto a high-frequency AC square wave.

The most predominant method for rendering fingertip cutaneous haptic in kinesthetic haptic devices involves incorporating additional vibration actuators, such as voice coils [3][35]. This novel HASEL actuator ability to simultaneously produce kinesthetic haptic feedback and haptic vibrations, highlights its unique advantages for haptic applications. However, we have observed a slight discrepancy between the frequency of the superimposed high-voltage sine waveform and the actuator vibration frequency. This difference increases as the frequency of the sine wave increases. Therefore, achieving precise control over vibration frequency is a future research direction to explore.

Other limitations of our HASEL actuator-based kinesthetic haptic device are: first, our proposed kinematic model of the force amplification and transmission mechanism in the device does not fit well with the data obtained from actual experiments (see Fig. 3). This prevents us from realizing model-based control. Second, we use load cell to measure real-time feedback force in our experiments to demonstrate the interactive performance of the device.

## 4. Conclusion

In this work, we designed a kinesthetic haptic feedback device based on a novel soft electrohydraulic actuator driven by high-frequency AC high voltage. Additionally, we developed a mathematical model and experimental validation demonstrated the device's capability to achieve stable and controllable feedback force output. Our experiments have demonstrated the device's capability for rapid force feedback output (high response speed). When under closed-loop control, the device achieved controllable fine feedback force output. Furthermore, by modulating the waveform of the driving high voltage, the device can generate variable-frequency haptic vibration output. We have further demonstrated the practical robustness of this kinesthetic haptic device by incorporating it into a teleoperated robotic system for real-world application scenarios. Future work will focus on the precision control of the vibration frequency as well as additional human subject experiments to better investigate the actual interaction perception of the user.

ACKNOWLEDGMENT

The authors would like to express their gratitude to JW Ambrose for manuscript writing assistance. This work was funded by the A*STAR Industry Alignment Fund - Pre-Positioning (A20H8A0241).

REFERENCES

[1] S. J. Lederman and R. L. Klatzky, "Haptic perception: A tutorial," *Attention Perception Psychophys.*, vol. 71, no. 7, pp. 1439–1459, 2009.

[2] K. Salisbury, F. Conti, and F. Barbagli, "Haptic Rendering: Introductory Concepts," *IEEE Computer Graphics and Applications*, vol. 24, no. 2, pp. 24-32, Mar.-Apr. 2004.

[3] I. Choi, H. Culbertson, M. R. Miller, A. Olwal, and S. Follmer, "Grabity:A wearable haptic interface for simulating weight and grasping in virtualreality," in *Proc. 30th Annu. ACM Symp. User Interface Softw. Technol.*,2017, pp. 119–130.

[4] I. Choi, E. Ofek, H. Benko, M. Sinclair, and C. Holz, "Claw: A multifunc-tional handheld haptic controller for grasping, touching, and triggering invirtual reality," in *Proc. CHI Conf. Hum. Factors Comput. Syst.*, 2018,pp. 1–13.

[5] I. Choi, E. W. Hawkes, D. L. Christensen, C. J. Ploch, and S. Follmer,"Wolverine: A wearable haptic interface for grasping in virtual reality," in *Proc. IEEE/RSJ Int. Conf. Intell. Robots Syst.*, 2016, pp. 986–993.

[6] R. Hinchet, V. Vechev, H. Shea, and O. Hilliges, "Dextres: Wearable hapticfeedback for grasping in vr via a thin form-factor electrostatic brake,"in *Proc. 31st Annu. ACM Symp. User Interface Softw. Technol.*, 2018,pp. 901–912.

[7] Q. Xiong, X. Liang, D. Wei, H. Wang, R. Zhu, T. Wang, J. Mao and H. Wang, "So-EAGlove: VR Haptic Glove Rendering Softness Sensation With Force-Tunable Electrostatic Adhesive Brakes," in *IEEE Transactions on Robotics*, vol. 38, no. 6, pp. 3450-3462, Dec. 2022, doi: 10.1109/TRO.2022.3172498.

[8] A. Bolopion and S. Régnier, "A Review of Haptic Feedback Teleoperation Systems for Micromanipulation and Microassembly," in *IEEE Transactions on Automation Science and Engineering*, vol. 10, no. 3, pp. 496-502, July 2013, doi: 10.1109/TASE.2013.2245122.

[9] Jacques Foottit, Dave Brown, Stefan Marks, and Andy M. Connor. "An Intuitive Tangible Game Controller," *In Proceedings of the 2014 Conference on Interactive Entertainment (IE2014)*. Association for Computing Machinery, New York, NY, USA, 1–7, doi: 10.1145/2677758.2677774.

[10] R. M. Pierce, E. A. Fedalei and K. J. Kuchenbecker, "A wearable device for controlling a robot gripper with fingertip contact, pressure, vibrotactile, and grip force feedback," *2014 IEEE Haptics Symposium (HAPTICS)*, Houston, TX, USA, 2014, pp. 19-25, doi: 10.1109/HAPTICS.2014.6775428.

[11] M. Cempini, M. Cortese and N. Vitiello, "A Powered Finger–Thumb Wearable Hand Exoskeleton With Self-Aligning Joint Axes," in *IEEE/ASME Transactions on Mechatronics*, vol. 20, no. 2, pp. 705-716, April 2015, doi: 10.1109/TMECH.2014.2315528.

[12] I. Zubrycki and G. Granosik, "Novel haptic device using jamming principle for providing kinaesthetic feedback in glove-based control interface," *Journal of Intelligent & Robotic Systems*, pp. 1–17, 2016.

[13] S. Jadhav, V. Kannanda, B. Kang, M. T. Tolley, and J. P. Schulze, "Soft robotic glove for kinesthetic haptic feedback in virtual reality environments," *Electron. Imag.*, vol. 3, pp. 19–24, 2017.

[14] Q. Xiong, X. Zhou, D. Li, J. W. Ambrose, R. C.-H. Yeow, "An Amphibious Fully-Soft Centimeter-Scale Miniature Crawling Robot Powered by Electrohydraulic Fluid Kinetic Energy." *Adv. Sci.* 2024, 11, 2308033, doi: 10.1002/advs.202308033.

[15] E. Acome et al. "Hydraulically amplified self-healing electrostatic actuators with muscle-like performance". In: *Science* 359.6371 (2018), pp. 61–65.

[16] N. Kellaris, V. G. Venkata, G. M. Smith, S. K. Mitchell, and C. Keplinger, "Peano-HASEL actuators: Muscle-mimetic, electrohydraulic transducers that linearly contract on activation," Sci. Robot., vol. 3, no. 14, 2018, doi: 10.1126/scirobotics.aar3276.

[17] E. Leroy, R. Hinchet, and H. Shea, "Multimode hydraulically amplified electrostatic actuators for wearable haptics," *Adv. Mater.*, vol. 32, no. 36,2020, Art. no. 2002564.

[18] P. Rothemund, N. Kellaris, S. K. Mitchell, E. Acome, and C. Keplinger, "HASEL artificial muscles for a new generation of lifelike robots— recent progress and future opportunities," *Advanced Materials*, vol. 33, no. 19, 2021.

[19] N. Kellaris et al., "Spider-inspired electrohydraulic actuators for fast, softactuated joints," *Adv. Sci.*, vol. 8, no. 14, May 2021, Art. no. 2100916.

[20] Z. Yoder, D. Macari, G. Kleinwaks, I. Schmidt, E. Acome, and C. Keplinger, "A Soft, Fast and Versatile Electrohydraulic Gripper with Capacitive Object Size Detection," *Adv. Funct. Mater.*, vol. 33, no. 3, 2023, doi: 10.1002/adfm.202209080.

[21] T. Nakamura and A. Yamamoto, "Modeling and control of electroadhesion force in DC voltage," *Robomech J.*, vol. 4, no. 1, pp. 1–10, 2017, doi: 10.1186/s40648-017-0085-3.

[22] F. Giraud, M. Amberg and B. Lemaire-Semail, "Merging two tactile stimulation principles: electrovibration and squeeze film effect," *2013 World Haptics Conference (WHC)*, Daejeon, Korea (South), 2013, pp. 199-203, doi: 10.1109/WHC.2013.6548408.

[23] Z. Yoder et al., "Design of a high-speed prosthetic finger driven by Peano-HASEL actuators," *Front. Robot. AI*, vol. 7, 2020, Art. no. 586216.

[24] S. D. Gravert et al., "Low-voltage electrohydraulic actuators for untethered robotics," *Sci. Adv.*, vol. 10, no. 1, Jan. 2024, Art. no. eadi9319.

[25] S. Kim and Y. Cha, "Electrohydraulic actuator based on multiple pouch modules for bending and twisting," *Sensors Actuators A Phys.*, vol. 337, no. February, p. 113450, 2022, doi: 10.1016/j.sna.2022.113450.

[26] Xiong, Q., Zhou, X., Li, D. and Yeow, R.C.H., 2024. "AC-Driven Series Elastic Electrohydraulic Actuator for Stable and Smooth Displacement Output," *arXiv preprint arXiv:2401.13941*.

[27] K. Qian, A. Song, J. Bao, and H. Zhang, "Small teleoperated robot for nuclear radiation and chemical leak detection," *International Journal of Advanced Robotic Systems*, vol. 9, no. 3, p. 70, 2012.

[28] P. Desbats, F. Geffard, G. Piolain, and A. Coudray, "Force-feedback teleoperation of an industrial robot in a nuclear spent fuel reprocessing plant", *Industrial Robot*, Vol. 33 No. 3, pp. 178-186, doi: 10.1108/0143991061070300.

[29] J. P. Clark, G. Lentini, F. Barontini, M. G. Catalano, M. Bianchi and M. K. O'Malley, "On the role of wearable haptics for force feedback in teleimpedance control for dual-arm robotic teleoperation," *2019 International Conference on Robotics and Automation (ICRA)*, Montreal, QC, Canada, 2019, pp. 5187-5193, doi: 10.1109/ICRA.2019.8793652.

[30] A. Okamura, "Methods for Haptic Feedback in Teleoperated Robot-Assisted Surgery," *Industrial Robot: An Int'l J.*, vol. 31, no. 6, pp. 499-508, 2004.

[31] C. Yang, J. Luo, C. Liu, M. Li and S. -L. Dai, "Haptics Electromyography Perception and Learning Enhanced Intelligence for Teleoperated Robot," in *IEEE Transactions on Automation Science and Engineering*, vol. 16, no. 4, pp. 1512-1521, Oct. 2019, doi: 10.1109/TASE.2018.2874454.

[32] I. Sarakoglou, N. Garcia-Hernandez, N. G. Tsagarakis and D. G. Caldwell, "A High Performance Tactile Feedback Display and Its Integration in Teleoperation," in *IEEE Transactions on Haptics*, vol. 5, no. 3, pp. 252-263, Third Quarter 2012, doi: 10.1109/TOH.2012.20.

[33] R. V. Patel, S. F. Atashzar and M. Tavakoli, "Haptic Feedback and Force-Based Teleoperation in Surgical Robotics," in *Proceedings of the IEEE*, vol. 110, no. 7, pp. 1012-1027, July 2022, doi: 10.1109/JPROC.2022.3180052.

[34] D. F. Glas, T. Kanda, H. Ishiguro and N. Hagita, "Teleoperation of Multiple Social Robots," in *IEEE Transactions on Systems, Man, and Cybernetics - Part A: Systems and Humans*, vol. 42, no. 3, pp. 530-544, May 2012, doi: 10.1109/TSMCA.2011.2164243.+

[35] J. Rekimoto, "Traxion: a tactile interaction device with virtual force sensation," in *Symposium on User Interface Software and Technology*, 2013, pp. 427–431.